\documentclass[journal,comsoc]{IEEEtran}
\usepackage[T1]{fontenc}

\ifCLASSINFOpdf
\else
\fi
\usepackage{amsmath}
\usepackage[cmintegrals]{newtxmath}

\usepackage{times}
\usepackage{soul}
\usepackage{url}
\usepackage[hidelinks]{hyperref}
\usepackage[utf8]{inputenc}
\usepackage{graphicx}
\usepackage{subcaption}
\usepackage{amsmath}
\usepackage{booktabs}
\usepackage{algorithm}
\usepackage{algorithmic}
\usepackage{pifont}
\usepackage{bbm}
\usepackage{cite}
\usepackage{color}
\usepackage{multirow}
\usepackage{multicol}

\begin{document}
%
\title{DreamNet: A Deep Riemannian Network based on SPD Manifold Learning for Visual Classification}
%
%
%

\author{Rui Wang,
        Xiao-Jun Wu*,
        Ziheng Chen,
        Tianyang Xu,
        and~Josef Kittler,~\IEEEmembership{Life~Member,~IEEE}
\thanks{R. Wang, X.-J. Wu (\textit{Corresponding author}), Z. Chen, and T. Xu are with the School of Artificial Intelligence and Computer Science, Jiangnan University, Wuxi 214122, China. R. Wang, X.-J. Wu, Z. Chen, and T. Xu are also with Jiangsu Provincial Engineering Laboratory of Pattern Recognition and Computational Intelligence, Jiangnan University. E-mail: (cs\_wr@jiangnan.edu.cn; xiaojun\_wu\_jnu@163.com; zh\_chen@stu.jiangnan.edu.cn; tianyang\_xu@163.com).}
\thanks{J. Kittler is with the Centre for Vision, Speech and Signal Processing,
University of Surrey, Guildford GU2 7XH, U.K. J. Kittler is also with the School of Artificial Intelligence and Computer Science, Jiangnan University. E-mail: (j.kittler@surrey.ac.uk).}
}

\markboth{Journal of \LaTeX\ Class Files.}%
{Shell \MakeLowercase{\textit{et al.}}: Bare Demo of IEEEtran.cls for IEEE Communications Society Journals}

\maketitle

\begin{abstract}
Image set-based visual classification methods have achieved remarkable performance, via characterising the image set in terms of a non-singular covariance matrix on a symmetric positive definite (SPD) manifold.
To adapt to complicated visual scenarios better, several Riemannian networks (RiemNets) for SPD matrix nonlinear processing have recently been studied.
However, it is pertinent to ask, whether greater accuracy gains can be achieved by simply increasing the depth of RiemNets. 
The answer appears to be negative, as deeper RiemNets tend to lose generalization ability. 
To explore a possible solution to this issue,  we propose a new architecture for SPD matrix learning. 
Specifically, to enrich the deep representations, we adopt SPDNet \cite{spdnet} as the backbone, with a stacked Riemannian autoencoder (SRAE) built on the tail. 
The associated reconstruction error term can make the embedding functions of both SRAE and of each RAE an approximate identity mapping, which helps to prevent the degradation of statistical information. 
We then insert several residual-like blocks with shortcut connections to augment the representational capacity of SRAE, and to simplify the training of a deeper network. 
The experimental evidence demonstrates that our DreamNet 
can achieve improved accuracy with increased depth of the network.
\end{abstract}


\IEEEpeerreviewmaketitle

\section{Introduction}

\IEEEPARstart{T}{he} advancement of multimedia technology makes video generation ubiquitous. However, the technology for the extraction of information from the videos is lagging behind. The problem is challenging not only because of the volume of data, but also because thanks to the changes in natural factors such as morphology, illumination, occlusion, and viewpoint, any recorded video data will exhibit a wide range of variations in scenes and objects. This calls for video-based computer vision and pattern recognition methods, including clustering and classification, that can work in less restricted conditions \cite{kim,zhou}. As each video sequence can be naturally converted into an image set data, in recent years, image set-based visual classification has attracted considerable attention, with reports of remarkable progress in a variety of practical scenarios, such as dynamic scene classification \cite{dmk,gepml,symnet}, video-based facial emotional recognition \cite{spdnet,gemkml,spdnetbn}, video-based face recognition \cite{leml,dml,slbf,spdml,dan}, and action recognition \cite{ps3d,jdrml,hgrnet,spdnetrp}. In such tasks, both the training and test samples are image sets, each of which contains a number of image samples of the same visual content.
Compared to the conventional, single still image-based classification, features from image sets are more informative, helping to enhance the probability of the correct interpretation of the image data.
Nevertheless, it is still unclear how best to characterize image sets and how to capture the variety of information they provide.

With the capacity to capture and characterize the spatiotemporal variations of data conveyed by image sequences \cite{cdl,leml,spdnet,spdml,spdnetrp}, covariance matrix has attracted a widespread attention in image set representation. However, the difficulty of processing and classifying such SPD matrices is that their underlying space is a curved Riemannian manifold, \textit{i.e.}, an SPD manifold \cite{ArsV}. Consequently, the tools applicable to Euclidean geometry cannot directly be used for computation. Thanks to the well-studied Riemannian metrics, including Log-Euclidean Metric (LEM) \cite{ArsV} and Affine-Invariant Riemannian Metric (AIRM) \cite{PenX}, the Euclidean methods can be generalized to the SPD manifold by utilizing the Riemannian kernel functions for mapping it into a Reproducing Kernel Hilbert Space (RKHS) \cite{cdl,vem,rsr,rcdl,mrmml,hrgeml}.
However, this type of approach may lead to undesirable solutions as it distorts the geometrical structure of the data manifold. To respect the original Riemannian geometry more faithfully, several recently suggested geometry-aware metric learning algorithms \cite{leml,cml,spdml,ps3d} have been developed to find a manifold-to-manifold embedding mapping, such that a discriminative space with the same geometry can be constructed.
Regrettably, despite their notable success, the existing  shallow linear transformation schemes for SPD matrices, implemented on nonlinear manifolds, impede these methods from mining fine-grained geometric representations.

\begin{figure}[!t]
 \centering
 \includegraphics[scale=0.57]{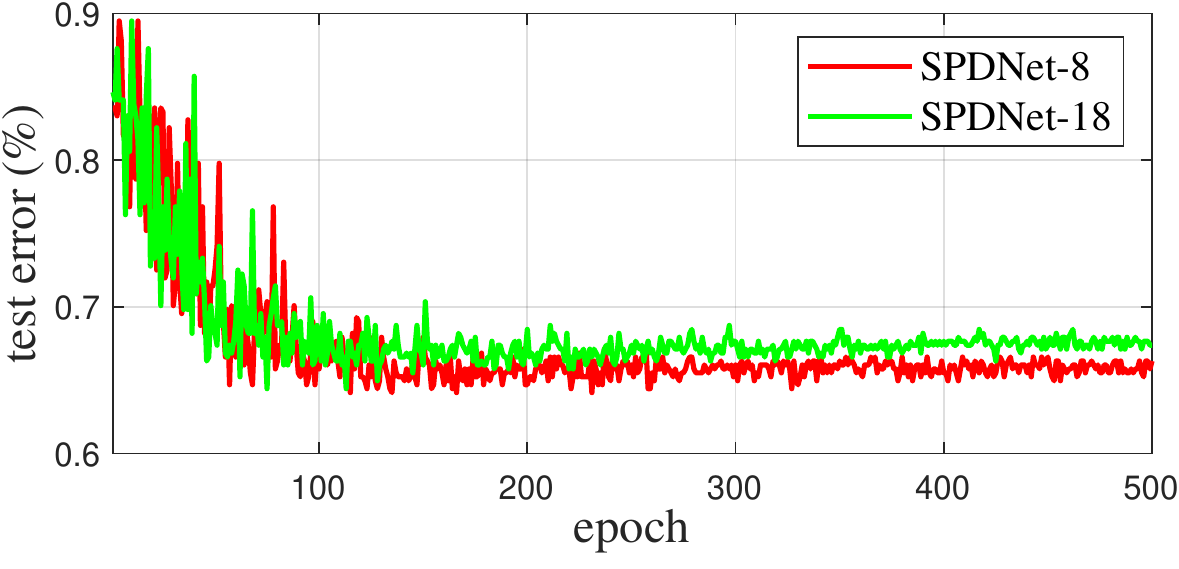}
 \vskip -0.07in
 \caption{The classification error of 8-layer SPDNet and 18-layer SPDNet versus the number of epochs on the AFEW dataset}  
 \label{fig-1}
 \vskip -0.10in
\end{figure}

Motivated by the powerful feature learning capability of convolutional neural networks (ConvNets) \cite{resnet,vgg}, an end-to-end Riemannian architecture for SPD matrix nonlinear learning has recently been proposed (SPDNet \cite{spdnet}).
The structure of SPDNet is constituted by a number of trainable blocks, each of which contains an SPD matrix bilinear mapping layer and an activation layer for data transformation and regularization. 
Subsequently, a Riemannian-Euclidean mapping layer maps the learned feature manifolds into a flat space for classification. 
More architectures have followed thereafter \cite{spdnetbn,symnet,hgrnet,dmtnet,spdnetrp,msnet}, modifying the elementary building blocks for different application scenarios. Compared to the aforementioned SPD matrix learning methods, the strength of this type of approach lies in the generalization of a shallow linear feature embedding scheme to a deep and nonlinear function, capable of learning deep representations with improved discriminability for better understanding of visual scenes. As recent evidence \cite{vgg,resnet} reveals, the network depth is of vital importance for promoting good performance. A question therefore arises: \textit{can the classification accuracy be improved by simply stacking more layers on top of each other in the SPD neural networks?} The following three factors make it impossible to provide ready  answers: 1) existing RiemNets have a small number of layers and there is no prior experience in building very deep networks; 2) there is limited research on this topic; 3) deeper SPD network exhibit a loss of generalisation capability. 
A typical example is illustrated in Fig. \ref{fig-1}. It should be noted that the classification error of SPDNet-18 is higher than that of SPDNet-8. 

\begin{figure*}[!t]
 \centering
 \includegraphics[scale=0.47]{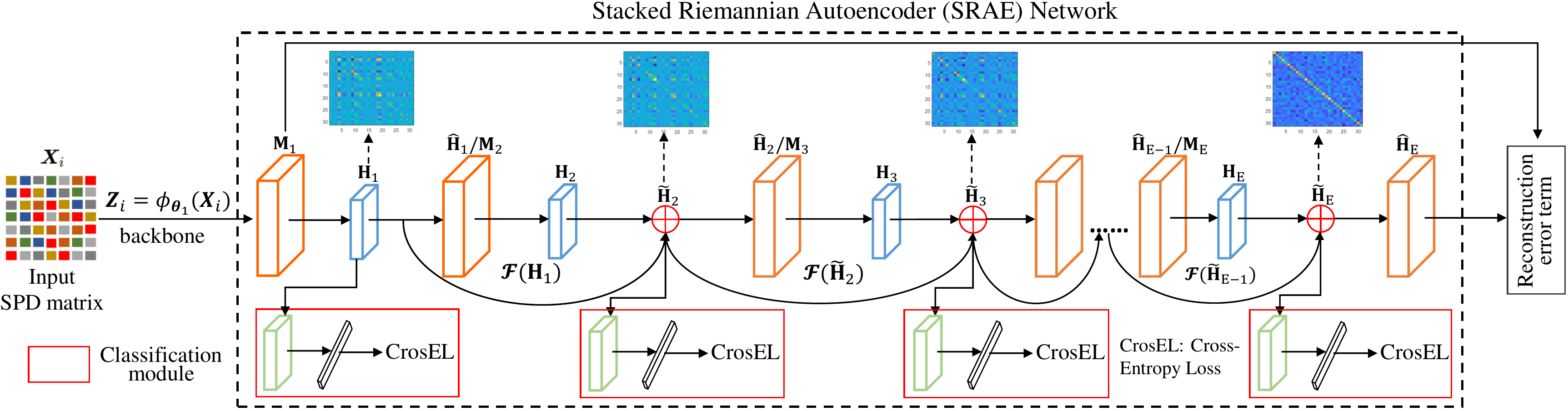}
 \vskip -0.05in
 \caption{A schematic diagram of the proposed DreamNet. 
In this figure, $\boldsymbol{\mathcal{F}}$ denotes the to-be-learnt Riemannian residual mapping.} 
 \label{fig-2}
 \vskip -0.05in
\end{figure*}

The above observation suggests that simply stacking more layers on top of each other does not mean that a better RiemNet can be learnt. This article proposes a new architecture for SPD matrix processing and classification that avoids the pitfalls of layer stacking in RiemNet. The overall framework of our approach is shown in Fig. \ref{fig-2}. As a greater depth of representation is essential for many visual classification tasks, the purpose of the proposed network is to learn a deeper manifold-to-manifold embedding that would transform the input SPD matrices into more informative ones of lower dimensionality and the same Riemannian geometry. To meet this requirement, 
we select SPDNet \cite{spdnet} as the backbone of the suggested model, in view of its demonstrable strength in SPD matrix nonlinear learning. Then, a stacked Riemannian autoencoder network (SRAE) is appended at the end of the backbone to increase the depth of the structured representations. Under the supervision of a reconstruction error term associated with the input-output SPD matrices of SRAE, the embedding mechanisms guide both SRAE and each RAE to approach an identity mapping, thus enabling them to prevent a degradation of the statistical information of the deeper features. The proposed solution ensures that the classification error produced by our deeper model will not be higher than that of the shallower backbone. To enhance the capacity of SRAE, we build multiple residual-like blocks within it, implemented by the shortcut connections \cite{resnet} between the hidden layers of any two adjecent RAEs. This design makes the current RAE learning stage access the informative features of the previous stages easily, thus facilitating the reconstruction of the remaining structural details. 

Since the above design ensures that the SRAE network remains sensitive to the data variations in the new feature manifolds, we also append a classification module, composed of the LogEig layer, FC layer, and cross-entropy loss to each RAE to facilitate manifold-to-manifold deep transformation learning. In this manner, a series of effective classifiers is obtained. 
Finally, a simple maximum voting strategy is applied for the final image set classification. 

We demonstrate the benefits of the proposed approach on the tasks of video-based facial emotion recognition, skeleton-based hand action recognition, and skeleton-based action recognition with UAVs, respectively. The experiments on the three benchmarking datasets show that our DreamNet achieves accuracy gains from an increasing network depth, producing better results than the previous methods.

\section{Related Works}
To generalize the shallow linear feature embedding mechanism of the conventional SPD matrix discriminant learning methods to deep and nonlinear function, Ionescu et al. \cite{d2p} integrate global Riemannian computation layers for processing SPD matrices into deep networks to capture structured features for scene understanding. In order to mine the geometric information of the original visual data embodied in the SPD matrices, Huang et al. \cite{spdnet} design a novel Riemannian neural network for SPD matrix nonlinear learning, comprising of a stack of SPD matrix transformation and activation layers, referred to as SPDNet. 
To provide a better guidance for the network learning, Brooks et al. \cite{spdnetbn} design a Riemannian batch normalization module for SPDNet.
To make a better use of the local structural information of the SPD matrix, Zhang et al. \cite{dmtnet} propose an SPD matrix 2D convolutional layer for data transformation, requiring each convolutional kernel also to be SPD. \textcolor{black}{Different from \cite{dmtnet}, Chakraborty et al. \cite{manifoldnet} simulate the weighted sums of CNN by designing a novel deep network for manifold-valued data using the weighted Fréchet Mean (wFM) operation to realize 'convolutions' on the manifold.} 
More recently, Wang et al. \cite{symnet} design a lightweight cascaded neural network for SPD matrix nonlinear learning and classification, showing higher computational efficiency and competitive classification performance, especially with limited training data.

\section{Proposed Method}
Although the Riemannian neural network approaches to SPD matrix processing can alleviate the negative impact of the variations of visual representations on the classification performance, achieving accuracy gains is not simply a matter of increasing the network depth.  The main obstacle to this simplistic solution is 
the degradation of statistical information (degradation problem),
which makes the learned deep representations unable to effectively characterize the structural information of the original imaged scene, thus resulting in lower accuracy. In this paper, we design a novel architecture named DreamNet to solve this issue. Fig. \ref{fig-2} provides an overview of our approach.

\subsection{Preliminaries}
\textbf{SPD Manifold:} A real-valued matrix $\boldsymbol{X}$ is called SPD if and only if $\boldsymbol{v}^T\boldsymbol{Cv}>0$ for all non-zero vector $\boldsymbol{v}\in\mathbbm{R}^d$. The space spanned by a set of $d$-by-$d$ SPD matrices is a Lie group, with a manifold structure denoted as $\boldsymbol{\mathcal{S}}_{++}^d$. More formally:
\begin{equation}
\begin{split}
    \boldsymbol{\mathcal{S}}_{++}^d := \{\boldsymbol{X}\in\mathbbm{R}^{d\times d}:\,&\boldsymbol{X}=\boldsymbol{X}^T,\,\\
    &\boldsymbol{v}^T\boldsymbol{Xv}>0,\forall \boldsymbol{v}\in\mathbbm{R}^d\backslash\{0_d\}\}.
\end{split}
\end{equation}
This enables the use of concepts related to differential geometry to address $\boldsymbol{\mathcal{S}}_{++}^d$, such as geodesic.

\textbf{Set Modeling with Second-Order Statistics:} Let $\boldsymbol{S}_i=[\boldsymbol{s}_1,\boldsymbol{s}_2,...,\boldsymbol{s}_{n_i}]$ be a $i^{\rm{th}}$ given image set (video sequence) with $n_i$ entries, where $\boldsymbol{s}_j \in \mathbbm{R}^{d\times1}$ denotes the $j^{\rm{th}}$ vectorized frame. For $\boldsymbol{S}_i$, its corresponding second-order representation is computed by:
\vskip -0.05in
\begin{equation}
   \boldsymbol{{X}}_i=\frac{1}{n_i-1}\sum_{j=1}^{n_i}(\boldsymbol{s}_j-\boldsymbol{u}_i)(\boldsymbol{s}_j-\boldsymbol{u}_i)^T.
\label{e1}
\end{equation}
where $\boldsymbol{u}_i$ is the mean of $\boldsymbol{S}_i$, expressed as $\boldsymbol{u}_i=\frac{1}{n_i}\sum_{j=1}^{n_i}\boldsymbol{s}_j$. Considering that $\boldsymbol{X}_i$ does not necessarily satisfy the condition of positive definiteness, it is regularised, \textit{i.e.}, $\boldsymbol{X}_i \gets \boldsymbol{X}_i+\lambda \boldsymbol{\rm{I}}_d$, where $\boldsymbol{\rm{I}}_d$ is an identity matrix of size $d \times d$, and $\lambda$ is set to $trace( \boldsymbol{X}_i)\times 10^{-3}$ in all the experiments. In this way, $\boldsymbol{X}_i$ is a true SPD manifold-valued element \cite{PenX}.

\textbf{Basic Layers of the SPD Manifold Neural Network:} 
Let $\boldsymbol{X}_{k-1}\in\boldsymbol{\mathcal{S}}_{++}^{d_{k-1}}$ be the input SPD matrix of the $k^{\rm{th}}$ layer. The Riemannian operation layers defined in the original SPDNet \cite{spdnet} are as follows:

\textit{BiMap Layer}: This layer is analogous to the usual dense layer for SPD data, used to transform the input data points into a lower dimensional space by a bilinear mapping $f_b$, expressed as $\boldsymbol{X}_k=f_b^{(k)}(\boldsymbol{{W}}_k,\boldsymbol{X}_{k-1})=\boldsymbol{{W}}_k^T\boldsymbol{X}_{k-1}\boldsymbol{{W}}_k$, where $\boldsymbol{{W}}_k$ is the column full-rank transformation matrix with semi-orthogonality to be learnt during training. 

\textit{ReEig Layer}: This layer is similar to the classical ReLU layers, designed to inject nonlinearity in the SPDNet by modifying the small positive eigenvalues of each input SPD matrix with a nonlinear rectification function $f_r$, formulated as $\boldsymbol{X}_k=f_r^{(k)}(\boldsymbol{X}_{k-1})=\boldsymbol{U}_{k-1}\mathrm{max}(\epsilon\boldsymbol{I},\boldsymbol{\Sigma}_{k-1})\boldsymbol{U}_{k-1}^T$, where $\boldsymbol{X}_{k-1}=\boldsymbol{U}_{k-1}\boldsymbol{\Sigma}_{k-1}\boldsymbol{U}_{k-1}^T$ represents the eigenvalue decomposition, and $\epsilon$ is a small activation threshold.

\textit{LogEig Layer}: To make the Euclidean computations applicable to the output feature manifold, the final LogEig layer is designed to perform the following mathematical operation: $\boldsymbol{X}_k=f_l^{(k)}(\boldsymbol{X}_{k-1})=\boldsymbol{U}_{k-1}\mathrm{log}(\boldsymbol{\Sigma}_{k-1})\boldsymbol{U}_{k-1}^T$, where $\boldsymbol{X}_{k-1}=\boldsymbol{U}_{k-1}\boldsymbol{\Sigma}_{k-1}\boldsymbol{U}_{k-1}^T$ denotes the eigenvalue decomposition.
With this operation, the conventional FC layers can be utilized in the resulting flat space for visual classification tasks.

\subsection{Deep Riemannian Network}
As presented in Fig. \ref{fig-2}, the designed SRAE module contains a cascade of Riemannian autoencoders (RAEs) to achieve continuous incremental reconstruction learning, in which the output feature maps of each RAE are used as the input data points of the adjacent one. 
To enrich the information flow in the SRAE network, we augment the sequential connections between adjacent RAEs using the shortcut connections, so that the current RAE module can effectively mine the revelant structural information with the aid of the former prediction for a better reconstruction. 
The network structure of each RAE is composed of three components. The first part is an encoder module, made up of the input (BiMap), nonlinear activation (ReEig), and hidden (BiMap) layers for SPD matrix dimensionality reduction, while maintaining Riemannian geometry. The second part is the decoder module, mainly used for data reconstruction. Since it has a  mirror-symmetric structure with the encoder, the RAE is defined strictly in the context of SPD manifolds, and so is SRAE and the whole network. Moreover, each RAE also connects to a classification network with the layers of LogEig and FC, guided by the cross-entropy loss.

Let $\boldsymbol{\mathfrak{S}}=[\boldsymbol{S}_1,\boldsymbol{S}_2,...,\boldsymbol{S}_N]$ and $\boldsymbol{L}=[l_1,l_2,...,l_N]\in\mathbbm{R}^{1\times N}$ be the original image set data used for training and its corresponding label vector, respectively. 
Executing the operation defined in Eqn.(\ref{e1}) to process $\boldsymbol{S}_i\;(i=1\to N)$, we obtain a new training set comprised of $N$ SPD matrices, denoted by $\boldsymbol{\mathcal{X}}=[\boldsymbol{X}_1,\boldsymbol{X}_2,...,\boldsymbol{X}_N]$. 
For the $i^{\rm{th}}$ input $\boldsymbol{X}_i$ of our DreamNet, the low dimensional and compact output data representation of the backbone can be expressed as $\boldsymbol{\mathcal{Z}}_i=\phi_{\boldsymbol{\theta}_1}(\boldsymbol{X}_i)$. Here, $\phi_{\boldsymbol{\theta}_1}$ represents the Riemannian network embedding from the input data manifold to the target one, implemented by a stack of BiMap and ReEig layers. Besides, $\boldsymbol{\theta}_1$ indicates the parameters of this backbone network to be learnt. 
As the SRAE module consists of $\rm{E}$ RAEs, we use 
$\boldsymbol{\rm{M}}_e\,(\boldsymbol{\rm{M}}_e=\boldsymbol{\mathcal{Z}}_i\,{\rm{when}}\,e=1)$, $\boldsymbol{\rm{H}}_e$, and $\boldsymbol{\rm{\hat{H}}}_e$ to respectively represent the input, output of the hidden layer, 
and reconstruction of the input of the $e^{\rm{th}}$ ($e=1\to \rm{E}$) RAE. Thus, $\boldsymbol{\rm{H}}_e$ and $\boldsymbol{\rm{\hat{H}}}_e$ can be computed by:
\begin{alignat}{1}
    &\boldsymbol{\rm{H}_e}= f_{b_e}(\boldsymbol{W}_{e_1},\boldsymbol{\rm{M}}_e)=\boldsymbol{W}_{e_1}^T\boldsymbol{\rm{M}}_e\boldsymbol{W}_{e_1},\\
    &\boldsymbol{\rm{\hat{H}}}_e=f_{b_e}(\boldsymbol{W}_{e_2},\boldsymbol{\rm{{H}}}_e)=\boldsymbol{W}_{e_2}^T\boldsymbol{\rm{{H}}}_e\boldsymbol{W}_{e_2},\label{e2}
\end{alignat}
where $f_{b_e}$ and $\boldsymbol{W}_{e_1}\in\mathbbm{R}^{d_{e-1}\times d_e}$ ($d_e\leq d_{e-1}$), $\boldsymbol{W}_{e_2}\in\mathbbm{R}^{d_e\times d_{e-1}}$ represent the bilinear mapping function and the transformation matrices of the $e^{\rm{th}}$ RAE, respectively. Since ${\boldsymbol{\rm{M}}}_e$ is actually equivalent to $\boldsymbol{\rm{\hat{H}}}_{e-1}$, we replace $\boldsymbol{\rm{M}}_e$ with $\boldsymbol{\rm{\hat{H}}}_{e-1}$ in the following for clarity.

\textcolor{black}{Based on the constructed SRAE architecture, the shortcut connections (SCs) and element-wise addition enable the Riemannian residual learning to be adopted for every set of a few stacked layers. In this article, we define the building block shown in Fig. \ref{fig-2} as:
\begin{equation}
\begin{split}
\boldsymbol{\rm{\tilde{H}}}_e=\boldsymbol{\rm{H}}_e+\boldsymbol{\rm{\tilde{H}}}_{e-1}=\boldsymbol{\mathcal{F}}(\boldsymbol{\rm{\tilde{H}}}_{e-1},\{\boldsymbol{W}_i\})+\boldsymbol{\rm{\tilde{H}}}_{e-1},
\end{split}
\label{sc}
\end{equation}
where $\boldsymbol{\rm{\tilde{H}}}_{e-1}$ and $\boldsymbol{\rm{\tilde{H}}}_e$ respectively represent the input and output of the residual-like block, $e=3\to\rm{E}$ (when $e=2$, $\boldsymbol{\rm{\tilde{H}}}_{e-1}$ is replaced by $\boldsymbol{\rm{{H}}}_{e-1}$ in Eqn.(\ref{sc}),  $\boldsymbol{\rm{H}}_e$ becomes equal to $\boldsymbol{\tilde{\rm{H}}}_e$ in Eqn.(\ref{e2})), and $\boldsymbol{\mathcal{F}}(\boldsymbol{\rm{\tilde{H}}}_{e-1},\{\boldsymbol{W}_i\})$ denotes the to-be-learnt Riemannian residual mapping. For example, $\boldsymbol{\mathcal{F}}=\boldsymbol{W}_{3_1}^T r(\boldsymbol{W}_{2_2}^T\boldsymbol{\tilde{\rm{H}}}_2\boldsymbol{W}_{2_2})\boldsymbol{W}_{3_1}$ when $e$ is set to 3, in which $r$ signifies the ReEig operation. 
In what follows, another ReEig nonlinearity is applied to the generated $\boldsymbol{\rm{\tilde{H}}}_e$ (\textit{i.e.}, $\boldsymbol{\rm{\tilde{H}}}_e\leftarrow r(\boldsymbol{\rm{\tilde{H}}}_e)$).
}

\subsection{Objective Function}
Briefly speaking, our goal is to probe a discriminative deep Riemannian network embedding to transform the input SPD matrices into more efficient and compact ones for improved classification. 
Taking the challenge of statistical information degradation caused by increasing the network depth into account, 
we establish a cascaded RAE module at the end of the backbone to reconstruct the remaining structural details from the input stage-by-stage.
The built residual-like blocks facilitate the reconstruction of the remaining residual by SRAE.
In addition, minimizing the reconstruction error term enables SRAE to remain highly sensitive to the variability of representations in the generated new feature manifolds, rendering the classification terms
to be more effective in encoding and learning the multi-view feature distribution information. 
Accordingly, the loss function of the proposed method is formulated as:
\begin{equation}
\small
     \mathcal{L}(\theta_2,\phi;\boldsymbol{\mathcal{X}})=\sum_{e=1}^{\rm{E}}\sum_{i=1}^N\mathcal{L}_e(\boldsymbol{X}_i, l_i)+\lambda\sum_{i=1}^N\mathcal{L}_2(\boldsymbol{\mathcal{Z}}_i, \boldsymbol{\rm{\hat{H}}}_E),
\label{ls}
\end{equation}
where $\boldsymbol{\theta}_2=\{\boldsymbol{\theta}_1,\boldsymbol{W}_{e_1},\boldsymbol{W}_{e_2},\boldsymbol{\mathcal{P}}_e\}$ ($\boldsymbol{\mathcal{P}}_e$ represents the projection matrix of the FC layer of the $e^{\rm{th}}$ RAE) and $\lambda$ is the trade-off parameter. In the experiments, we assign a small value to $\lambda$ to fine-tune classification performance. Further discussions of the role of this parameter will be presented later.

The first term of Eqn.(\ref{ls}) is the cross-entropy loss used to minimize the classification error of the input-target pairs ($\boldsymbol{X}_i,l_i$) ($i=1\to N$), implemented with the aid of the LogEig and FC layers. Specifically,  $\mathcal{L}_e$ is given as:
\begin{equation}
     \mathcal{L}_e(\boldsymbol{X}_i,l_i)=-\sum_{t=1}^cr(l_i,t)\times {\rm{log}}\frac{e^{\boldsymbol{\mathcal{P}}_e^t\boldsymbol{V}_e}}{\sum_{\tau}e^{\boldsymbol{\mathcal{P}}_e^{\tau}\boldsymbol{V}_e}},
\end{equation}
where $\boldsymbol{V}_e$ denotes the vectorized form of $\boldsymbol{\rm{\tilde{H}}}_e$ ($\boldsymbol{\rm{H}}_e$, when $e=1$), $\boldsymbol{\mathcal{P}}_e^t$ signifies the $t^{\rm{th}}$ row of the projection matrix $\boldsymbol{\mathcal{P}}_e\in\mathbbm{R}^{c\times (d_e)^2}$, and $r(l_i,t)$ is an indicator function, where $r(l_i,t)=1$ if $l_i=t$, and 0 otherwise. 

The second term of Eqn.(\ref{ls}) is the reconstruction error term (RT) measuring the discrepancy between the input sample and its corresponding reconstruction, computed by:
\begin{equation}
\mathcal{L}_2(\boldsymbol{\mathcal{Z}}_i, \boldsymbol{\rm{\hat{H}}}_E)=||\boldsymbol{\mathcal{Z}}_i-\boldsymbol{\rm{\hat{H}}}_{\rm{E}}||_{\rm{F}}^2. \label{rt}
\end{equation}
It is evident that the Euclidean distance (EuD) is utilized to supersede LEM for similarity measurement in Eqn.(\ref{rt}). Our motivation for this replacement is two-fold: 1) matrix inversion can be shunned in the backpropagation process; 2) EuD can measure the
’statistical-level’ similarity between SPD samples intuitively. Besides, the experimental results reported in Table \ref{tab-rt} show that although the use of LEM can lead to a certain improvement in classification performance, the computation time requied is close to three times that of EuD, confirming the rationality of using EuD.
\begin{table}[!t] 
\renewcommand\arraystretch{1.1}
\centering
\caption{Comparison of DreamNet-27 on the AFEW dataset.}
\label{tab-rt}
\resizebox{\linewidth}{!}{
\begin{tabular}{l|c|c}
\hline
Metrics  &Acc. (\%) & Training time (s/epoch)\\
\hline
RT-EuD, \textit{i.e.}, Eq. (\ref{rt})  & 36.59 & 31.32 \\
RT-LEM  & 36.71 &  88.16\\
\hline
\end{tabular}}
\end{table}

\section{Experiments}
\label{s_exper}
In this section, we validate the efficacy of DreamNet\footnote{The source code will be available at: https://github.com/GitWR/DreamNet} on three typical visual classification tasks, namely video-based facial emotion recognition using the AFEW dataset \cite{afew}, skeleton-based hand action recognition using the FPHA dataset \cite{fpha}, and skeleton-based human action recognition using the UAV-Human dataset \cite{uav}, respectively. 

\subsection{Implementation}
To construct the network architecture of the backbone, we use five layers: $\boldsymbol{X}_i\to f_b^{(1)}\to f_{re}^{(2)}\to f_b^{(3)}\to f_{re}^{(4)}\to f_b^{(5)}$, where $f_b$ and $f_{re}$ denote the layers of BiMap and ReEig, respectively. The stacked Riemannian autoencoder network (SRAE) connected to the output of the backbone is formed by $\rm{E}$ RAEs, each of which can be grouped into two branch networks. The first branch is the RAE module,
making up five layers: $\boldsymbol{\hat{\mathrm{H}}}_{e-1}\to f_{b}$ (input) $\to f_{re}\to f_{b}$ (hidden) $\to f_{re}\to f_{b}$ (reconstruction). The second branch is the classification module, consisting of three layers: $\boldsymbol{\rm{\tilde{H}}}_e$ ($\boldsymbol{\rm{H}}_e$ when $e=1$)$\to f_{\mathrm{log}}\to f_{\mathrm{F}}\to f_{ce}$. 
Here, $f_{\mathrm{log}}$, $f_{\rm{F}}$, and $f_{ce}$ represent the LogEig layer, FC layer, and cross-entropy loss, respectively. 
In our experiments, the learning rate $\eta$ is set to 0.01, the batch size $B$ is  30, the weights of the BiMap and FC layers are initialized as random semi-orthogonal matrices and random matrices respectively, and the threshold $\epsilon$ of the ReEig layer is set to $1e$-4 for the AFEW and FPHA datasets and $1e$-5 for the UAV-Human dataset.
To train our DreamNet, we use an i7-9700 (3.4GHz) PC with 16GB RAM. We found that using GPU (GTX 2080Ti) does not speed up network training. The main bottleneck seems to be the series of eigenvalue operations.

\subsection{Dataset Description and Settings}
\textbf{AFEW Dataset:} 
This dataset consists of 2118 video clips (split in 1741+371 fixed training sets and validation sets) of natural facial expressions collected from movies. For the evaluation, we follow the protocols of \cite{spdnet,symnet} to scale down each video clip to a set of $20\times 20$ gray-scale images, such that a $400 \times 400$ SPD matrix can be computed for video (image set) representation. 
On this dataset, we set the filter sizes of the backbone to $400\times 200$, $200\times 100$, and $100\times 50$, and those of the $e^{\mathrm{th}}$ RAE are configured as $100\times 50$ and $50\times 100$.

\textbf{FPHA Dataset:} 
This dataset includes 1,175 hand action videos belonging to 45 different categories, performed by 6 actors in the first-person view. For the evaluation, we follow the criterion of \cite{fpha} to transfer each frame into a 63-dimensional vector using the 3D coordinates of 21 hand joints provided. Hence, a total of 1,175 SPD matrices of size $63\times 63$ can be computed to represent the data sequences, of which 600 are designated for training and the remaining 575 are used for testing. On this dataset, the sizes of the transformation matrices of the backbone are configured as $63\times 53$, $53\times 43$, and $43\times 33$, and those of the $e^{\mathrm{th}}$ RAE are set to $43\times 33$ and $33\times 43$. 

\textbf{UAV-Human:} 
This dataset is comprised of 22,476 video sequences representing 155 human action categories, collected by unmanned aerial vehicles (UAVs). As each pose frame is labeled by 17 major  body  joints with 3D coordinates, we follow the practice of \cite{czh-tbd} to shape each frame into a 51-dimensional feature vector. Considering that some actions are performed by two persons, we then use the PCA technique to transform the 102-dimensional vectors into 51-dimensional ones, by preserving 99\% energy of the data. In this scenario, each action video can be described by an SPD matrix of size $51\times 51$. Finally, the seventy-thirty-ratio (STR) protocol is applied to construct the gallery and probes from the randomly picked 16,724 SPD matrices. On this dataset, the sizes of the connection weights are set to ($51\times 43$, $43\times 37$, and $37\times 31$) and ($37\times 31$ and $31\times 37$) for the backbone and the $e^{\mathrm{th}}$ RAE, respectively.

\begin{figure}[!t]
 \centering
 \includegraphics[scale=0.54]{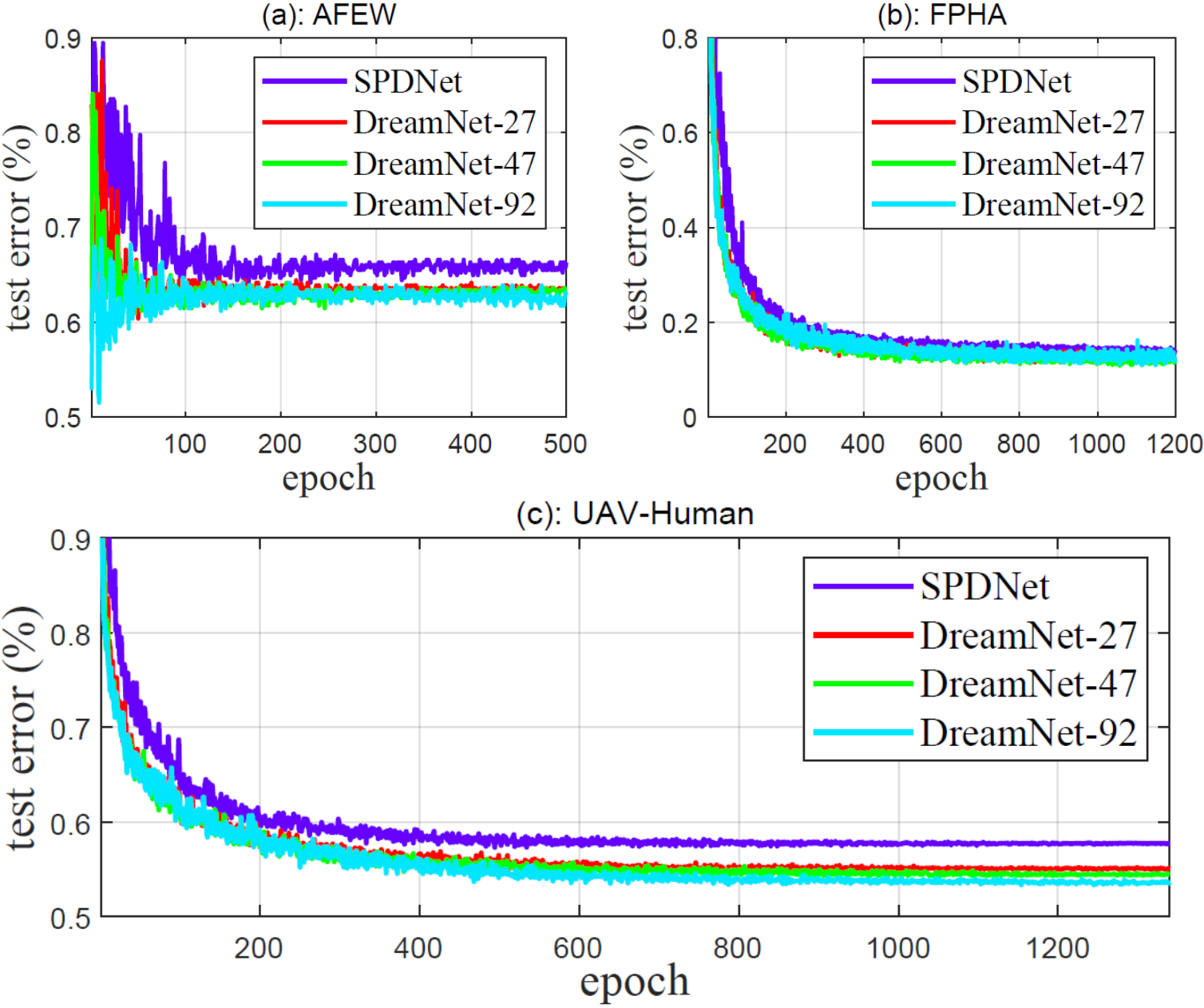}
 \caption{The classification error of the 27/47/97-layer DreamNets versus the number of training epochs on the AFEW, FPHA, and UAV-Human datasets.} 
 \label{fig-depth-1}
\end{figure}

\begin{figure}[!t]
 \centering
 \includegraphics[scale=0.37]{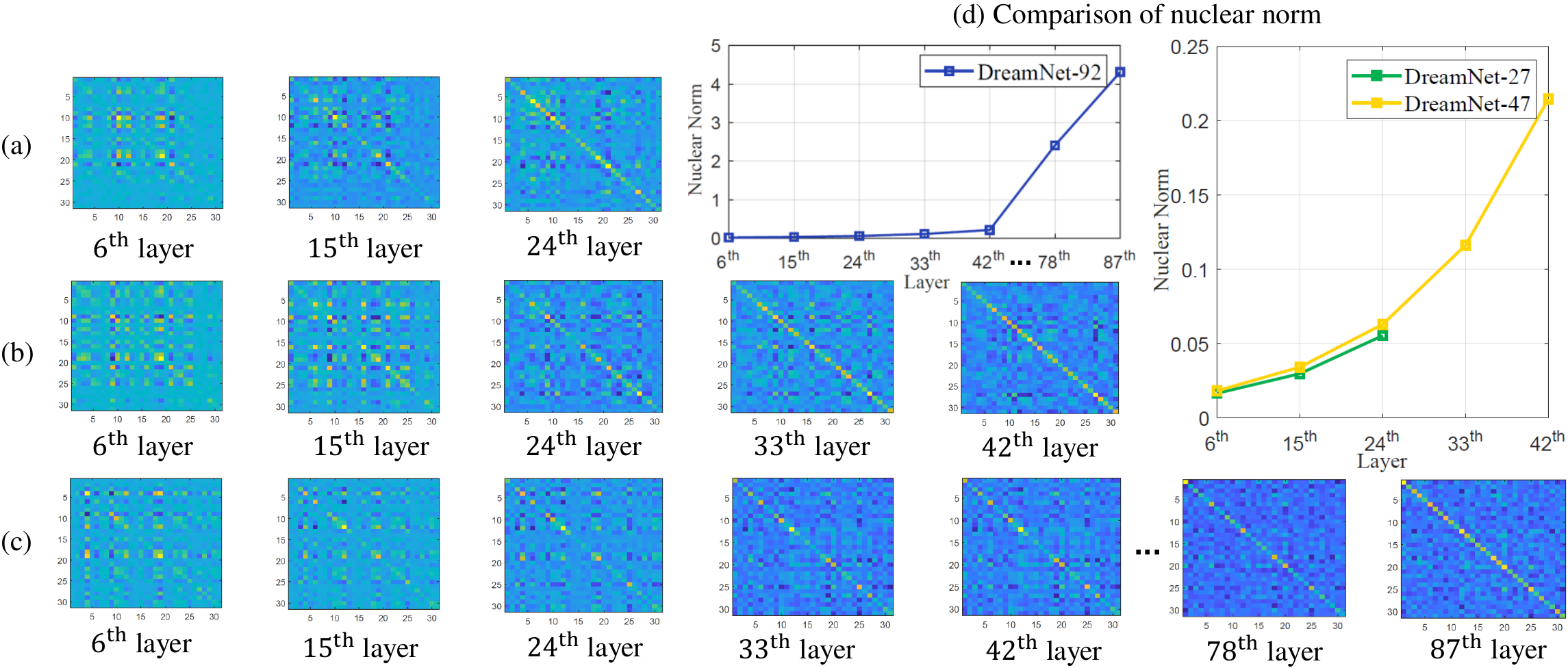}
 \caption{\textcolor{black}{The feature maps from different layers of the 27/47/92-layer DreamNets on the UAV-Human dataset are visualized in (a), (b), and (c), respectively. (d) shows the nuclear norms of these feature maps. Here, the $6^{\rm{th}}$ layer is actually the hidden layer of the first RAE, and the other layers are actually used to realize element-wise addition}.} 
 \label{fig-map}
\end{figure}

\subsection{Ablation Studies}
In this section, we perform experiments to study the effectiveness of the proposed approach in SPD matrix processing, learning, and image set-based visual classification. 

\textbf{Ablation for the DreamNet:} To evaluate our DreamNet, we carry out experiments on the AFEW, FPHA, and UAV-Human datasets to measure the impact of the network depth on the learning capacity of the proposed model. Based on the experimental results reported in Fig. \ref{fig-depth-1}(a), we can make three main observations. Firstly, the inverse correlation between the depth and network accuracy is reversed with the embedding function proposed in this paper, \textit{i.e.}, the 47-layer DreamNet ($\rm{E}=5$) performs better than the 27-layer DreamNet ($\rm{E}=3$). More importantly, the test error of DreamNet-47 is lower than that of DreamNet-27. This signifies that the degradation problem is conquered in this design and we succeed in improving accuracy with the increased depth. 
The consistency of these findings can be gleaned from Fig. \ref{fig-depth-1}(b) and Fig. \ref{fig-depth-1}(c). 

Secondly,  in the experiments, we also explore a 92-layer DreamNet by simply stacking more RAEs ($\rm{E}=10$ at this time). We find that compared with the 27/47-layer DreamNets, the 92-layer DreamNet achieves even lower test errors on the AFEW and UAV-Human datasets, demonstrating that the learning capacity of our network benefits from an extensive increase in the number of network layers. However, from Fig. \ref{fig-depth-1}(b) and Table \ref{tab-depth-2}, we note that the test error of DreamNet-92 is slightly higher than that of DreamNet-47 on the FPHA dataset. This could be caused by the relatively small size of this dataset.
Although the benefits of depth are reflected in the classification accuracy reported in Tables \ref{tab-depth-1}, \ref{tab-depth-2}, \ref{tab-depth-3}, the increase in network complexity (number of parameters and training speed) are detrimentally affected. 

Thirdly,  according to Fig. \ref{fig-depth-1}, we can see that the 27/47/92-layer DreamNets are easy to train on the three benchmarking datasets. The convergence speed of these three networks is greater than that of the original SPDNet. \textcolor{black}{Note that on the AFEW dataset, the test error of our 92-layer DreamNet first shows a degradation, but eventually it recovers and exhibits performance gains. We find that this behaviour is also mirrored by the loss function on the test set. The following two factors are the main reasons for overfitting: 1) the dataset contains only 7 categories and has large intra-subject ambiguity; 2) the network is too large.} 

\begin{table}[!t] 
\renewcommand\arraystretch{1.1}
\centering
\caption{Comparison on the AFEW dataset.}
\label{tab-depth-1}
\begin{tabular}{l|c|c|c}
\hline
Metrics  &DreamNet-27 & DreamNet-47 & DreamNet-92 \\
\hline
Acc.  & 36.59 & 36.98 & 37.47 \\
s/epoch  & 31.32 & 46.98 & 80.62 \\
\#params & 0.36M & 0.53M & 0.95M \\
\hline
\end{tabular}
\end{table}

\begin{table}[!t] 
\renewcommand\arraystretch{1.1}
\centering
\caption{Comparison on the FPHA dataset.}
\label{tab-depth-2}
\begin{tabular}{l|c|c|c}
\hline
Metrics  &DreamNet-27 & DreamNet-47 & DreamNet-92 \\
\hline
Acc. (\%)  & 87.78 & 88.64 & 88.12 \\
s/epoch  & 2.60 & 3.66 & 6.70 \\
\#params & 0.11M & 0.18M & 0.36M \\
\hline
\end{tabular}
\end{table}

\begin{table}[!t] 
\renewcommand\arraystretch{1.1}
\centering
\caption{Comparison on the UAV-Human dataset.}
\label{tab-depth-3}
\begin{tabular}{l|c|c|c}
\hline
Metrics  &DreamNet-27 & DreamNet-47 & DreamNet-92 \\
\hline
Acc. & 44.88 & 45.57 & 46.28 \\
s/epoch  & 49.04 & 71.33 & 129.29 \\
\#params & 0.10M & 0.16M & 0.31M \\
\hline
\end{tabular}
\end{table}

\textbf{Visualization:} To give the reader an intuitive feeling about the proposed method in addressing the degradation of structural information, we choose the UAV-Human dataset as an exmaple to visualize the SPD feature maps learned by the different layers of the 27/47/92-layer DreamNets. 
From Fig. \ref{fig-map}(a)-(c), we make two interesting observations: 1) for each DreamNet, compared to the low-level feature matrices, the magnitudes of the elements on the main diagonal of the high-level feature matrices are becoming larger, while the off diagonal ones are getting smaller; 2) with increasing the network depth, this concentration of energy becomes even more significant. Besides, the nuclear norms shown in Fig. \ref{fig-map}(d) reflect that the deeper the learned features, the lower their redundancy. These results demonstrate that the proposed method can effectively capture the feature areas with statistical positive and negative relevance in the original visual scene, thus being helpful for visual classification. 

\textcolor{black}{{Besides, the incremental information conveyed by the visualized feature maps demonstrate that the Riemannian residual mapping $\boldsymbol{\mathcal{F}}$ is not close to a zero mapping, which is different from the hypothesis verified in ResNet, \textit{i.e.}, the residual functions might be generally close to zero. The reason is that the semi-orthogonality of the weights makes it impossible for the Riemannian solver to drive the weights of the multiple layers towards zero. This shows that in the SPD network we studied, the Riemannian mapping obtained by a few stacked layers may not be close to an identity mapping}.}

\begin{figure}[!t]
 \centering
 \includegraphics[scale=0.48]{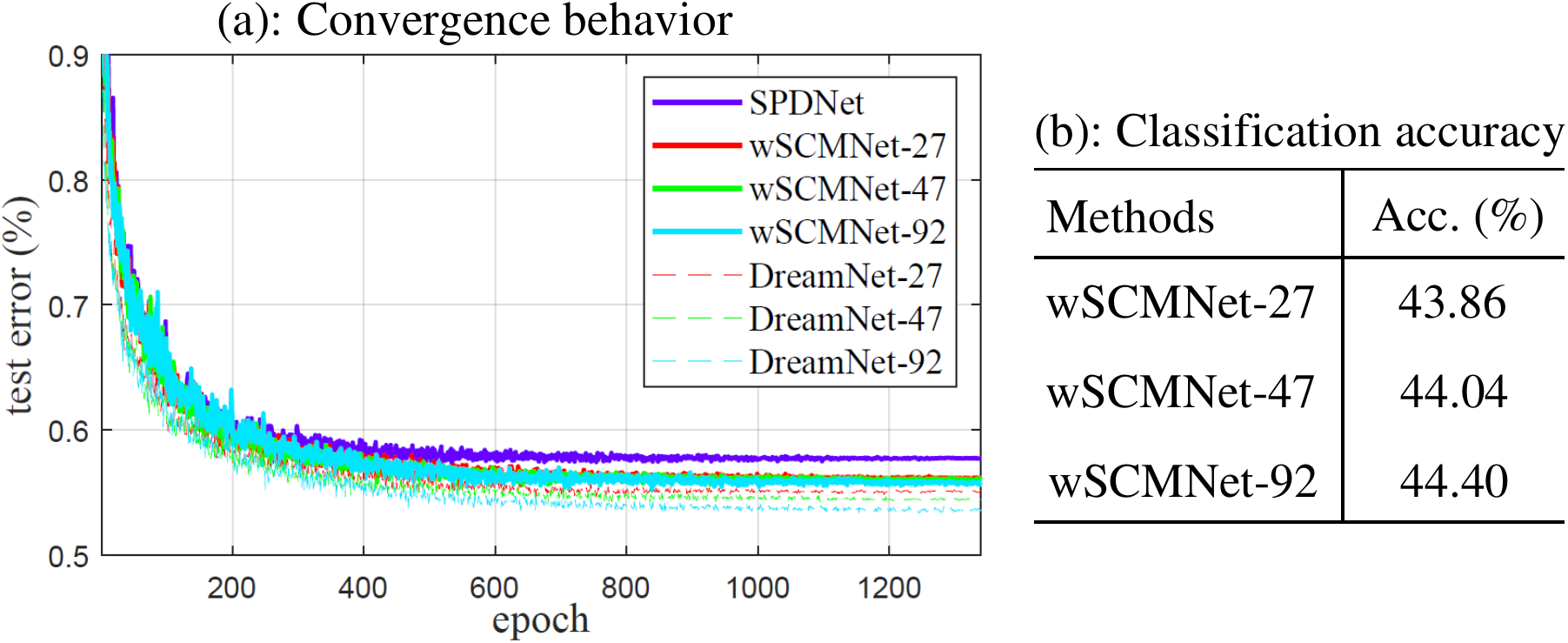}
 \caption{Performance on the UAV-Human dataset} 
 \label{fig-sc}
\end{figure}

\textbf{Ablation study for the Shortcut Connections:} To verify the benefits of the shortcut connections (SCs), we make experiments to study the performance of a simplified DreamNet (named wSCMNet) obtained by removing the SCs from SRAE module. We choose the UAV-Human dataset as an example. 
\textcolor{black}{It can be seen from Fig. \ref{fig-sc}(a) that the wSCMNet with different depths can converge to a better solution in less than 1,300 epochs, indicating that it has a good convergence behavior. However, the classification scores of 27/47/92-layer wSCMNets tabulated in Fig. \ref{fig-sc}(b) are lower than those of 27/47/92-layer DreamNets. In spite of this, they are still better than those of the competitors listed in Table \ref{tab-au}. From Fig. \ref{fig-sc}(a), we also find that the convergence speed of DreamNets is slightly faster than that of wSCMNets. \textcolor{black}{These experimental results not only demonstrate the effectiveness of the proposed SRAE network, but also confirm that the SCs: 1) enhance the capacity of SRAE module; 2) simplify the training of deeper networks. The underlying reason is that this operation facilitates the information interaction between different RAE learning stages.}}

\begin{figure}[!t]
 \centering
 \includegraphics[scale=0.50]{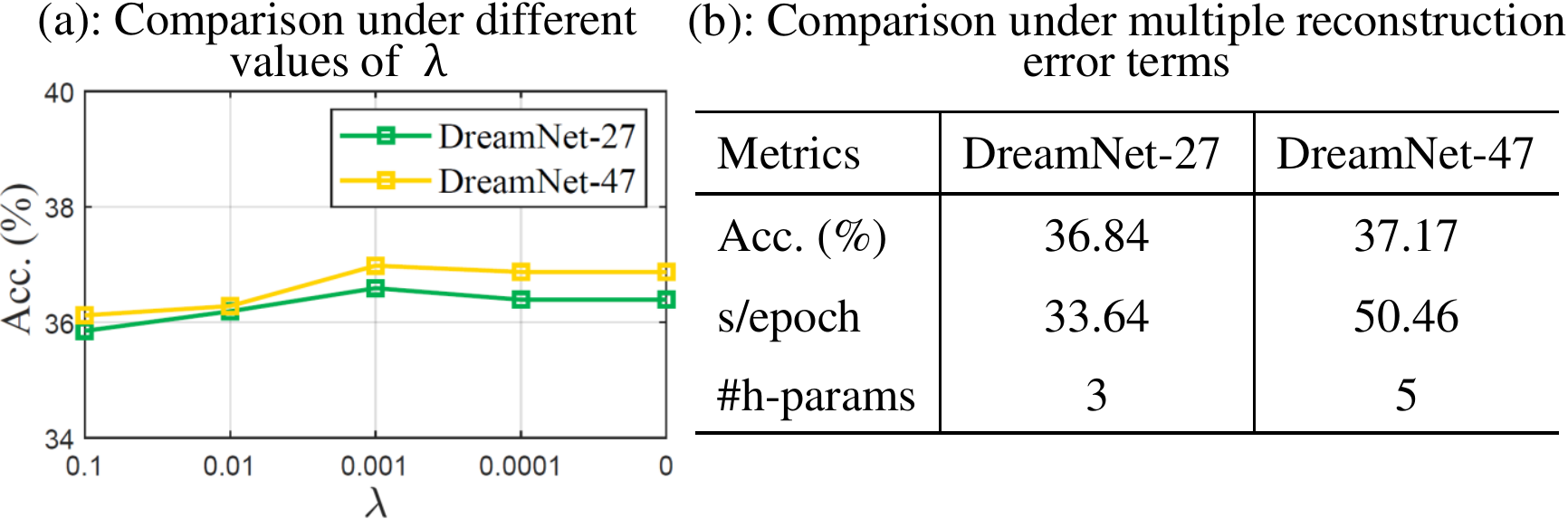}
 \caption{Comparison of DreamNet-27/47 on the AFEW dataset} 
 \label{fig-rt}
\end{figure}

\textbf{Ablation study of the role of the Reconstruction Error Term:} To measure the effectiveness of the reconstruction error term (RT), 
we select the AFEW dataset as an example to conduct the experiments. From Fig. \ref{fig-rt}(a), we have some interesting findings. Firstly, the performance of DreamNet-47 is better than that of DreamNet-27 in almost all cases, again demonstrating the benefits of increasing the network depth. 
Secondly, the classification accuracy of 27/47-layer DreamNets shows an increasing trend first and then decreasing.
This is mainly attributed to the fact that the loss function of our DreamNet has two goals: 1) supervising the network to generate deep representations with richly structured semantic information; 2) enabling the network  to reconstruct the input better. Note that  a large value of $\lambda$ would make the network focus on deep reconstruction learning, which is not conducive to the training of effective classifiers. 
\textcolor{black}{However, when $\lambda$ takes values in the range of \{0, 0.0001, 0.001\}, the performance of 27/47-layer DreamNets is slightly improved. 
This supports our assertion that the RT helps to fine-tune the classification performance. In any case, our method is not very sensitive to this trade-off parameter.} 

\textcolor{black}{Based on the above results, we connected all the input layers of the remaining $\rm{E}$-1 RAEs to the final layer of SRAE to include more RTs in DreamNet. Fig. \ref{fig-rt}(b) shows that this operation results in a relatively higher accuracy of 27/47-layer DreamNets, compared to Table \ref{tab-depth-1}. \textcolor{black}{The main reason is that the reconstruction learning at each layer of the network helps the high-level features capture the pivotal information variability conveyed by the original visual data, thus facilitating effective classification.} Meanwhile, the training time has not increased significantly. Nevertheless, as the number of hyperparameters (\#h-params) of our network increases, we suggest using just one RT shown in Fig. \ref{fig-2} as a compromise. 
All in all, these experiments confirm the efficacy of the RT in guiding our DreamNet to solve the degradation problem and to yield useful deep features.}

\begin{figure}[!t]
 \centering
 \includegraphics[scale=0.35]{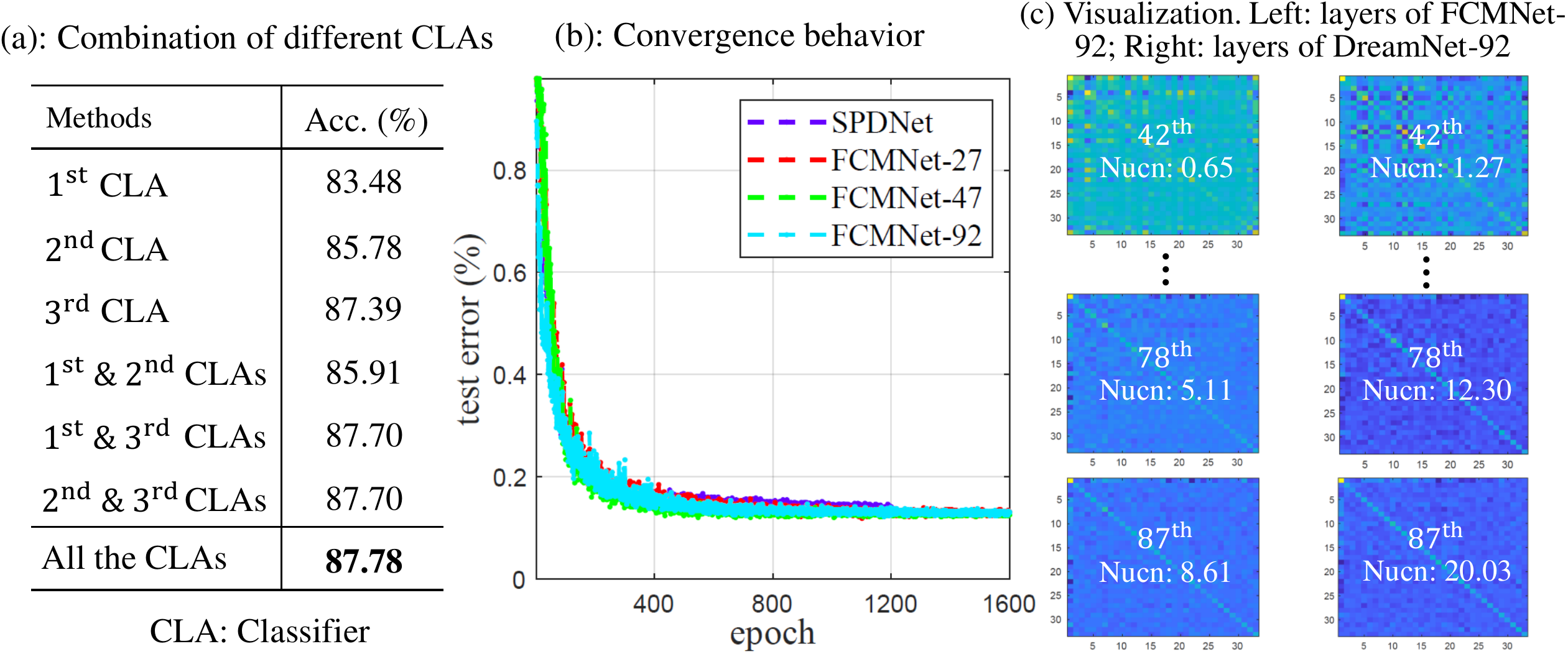}
 \caption{Performance  of FCMNet and DreamNet on the FPHA dataset, where 'Nucn' represents the nuclear norm.} 
 \label{fig-fcm}
\end{figure}

\textbf{Ablation of the Classification Module:}
\textcolor{black}{In this part, we make experiments on the FPHA dataset as an example to investigate the impact of the number of classification modules on the accuracy of DreamNet (here we take DreamNet-27 as an example) in the test phase. From Fig. \ref{fig-fcm}(a), we can see that: 1) the greater the number of classifiers, the higher the accuracy; 2) the $3^{\rm{rd}}$ classifier are more effective than the others. This not only indicates that these classifiers are complementary to each other, but also demonstrates that the higher-level features are more informative.} 

\textcolor{black}{Inspired by this experiment, we then investigated how the performance of DreamNet is affected by removing the first $\rm{E}$-1 classification modules from SRAE (we name the simplified DreamNet FCMNet here). In this case, we find that the initial learning rate of 0.01 is a bit too small for the 47/92-layer FCMNets. So we respectively assign the initial learning rates of 0.02 and 0.05 to FCMNet-47 and FCMNet-92, and make them attenuate by a factor of 0.9 every 100 epochs. It is evident that the studied FCMNets converge well (Fig. \ref{fig-fcm}(b)). Although the accuracy of 27/47/92-layer FCMNets (87.18\%, 87.60\%, and 87.30\%) are somewhat inferior to that of 27/47/92-layer DreamNets, they are still better than those of the competitors listed in Table \ref{tab-fpha}. These observations certify from another perspective that our design can overcome the degradation problem and yield a discriminative manifold-to-manifold deep transformation mapping for improved classification. \textcolor{black}{Besides, Fig. \ref{fig-fcm}(c) not only further indicates that the residual mapping $\boldsymbol{\mathcal{F}}$ is not approach to a zero mapping, but also shows that the multi-classifier learning (MCL) scheme can yield more efficient deep features with lower redundancy.} Since the use of multiple classification modules can provide sufficient supervision information, and the increase in training time is slight (\textit{e.g.}, one training epoch lasted on average 4.51s for FCMNet-92, and 6.70s for DreamNet-92 on this dataset), we adopt the MCL mechanism in this article.}

\begin{table}[!t]
\renewcommand\arraystretch{1.1}
\centering
\caption{Accuracy (\%)  comparison on the AFEW and UAV-Human datasets.} 
\label{tab-au}
\resizebox{\linewidth}{!}{
\begin{tabular}{l|c|c|c}
\hline
Methods & Source & AFEW & UAV-Human \\
\hline
GDA \cite{gda}   & ICML'08  & 29.11 & 28.13 \\
CDL \cite{cdl}   &  CVPR'12 & 31.81 & 31.11 \\
HERML \cite{herml}  & PR'15  & 32.14 &  N/A\\
MRMML \cite{mrmml} & TBD'22 & 35.71 & N/A \\
PML \cite{pml}   & CVPR'15 & 28.98 & 10.66\\
LEML \cite{leml} & ICML'15 & 25.13 & N/A\\
SPDML\cite{spdml} & TPAMI'18 & 26.72 & 22.69\\
GEMKML \cite{gemkml}  & TMM'21 & 35.71 & N/A\\
SymNet \cite{symnet} & TNNLS'22 & {32.70} &{35.89} \\
\textcolor{black}{ManifoldNet} \cite{manifoldnet} & TPAMI'20 & 23.98 & N/A\\
DeepO2P \cite{d2p} & ICCV'15 & 28.54 & N/A\\
\textcolor{black}{DARTS} \cite{darts} & ICLR'19 & 25.87 & \textcolor{black}{36.13} \\
\textcolor{black}{FairDARTS} \cite{faird} & ECCV'20 & 25.34 & \textcolor{black}{40.01} \\
GrNet \cite{grnet} & AAAI'18 & 34.23 &  35.23\\
SPDNet \cite{spdnet}  & AAAI'17 & 34.23 & 42.31\\
SPDNetBN \cite{spdnetbn}  & NeurIPS'19 & {36.12} & 43.28 \\
\hline
DreamNet-27 &  & {36.59} &  {44.88} \\
DreamNet-47  &  &  {36.98} &  {45.57}\\
DreamNet-92  &  &  \textbf{37.47} &  \textbf{46.28}\\
\hline
\end{tabular}}
\end{table}

\begin{table}[!t]
\renewcommand\arraystretch{1.1}
\centering
\caption{Accuracy (\%)  comparison on the FPHA dataset.}
\label{tab-fpha}
\resizebox{\linewidth}{!}{
\begin{tabular}{l|c|ccc|c}
\hline
Methods & Source & Color & Depth & Pose & Acc.  \\
\hline
Two streams \cite{ts} &CVPR'16 & \ding{51} & \ding{55} & \ding{55} & 75.30 \\
Novel View \cite{view}  & CVPR'16 & \ding{55}  & \ding{51}  & \ding{55} & 69.21 \\
Lie Group \cite{lie}  & CVPR'14 & \ding{55} & \ding{55} & \ding{51} & 82.69\\
HBRNN \cite{hbrnn}  & CVPR'15 & \ding{55} & \ding{55} & \ding{51} & 77.40 \\
LSTM \cite{fpha} & CVPR'18 & \ding{55} & \ding{55} & \ding{51} & 80.14 \\
JOULE\cite{joule} & CVPR'15 & \ding{51} & \ding{51} & \ding{51} & 78.78 \\
Gram Matrix \cite{gram} & CVPR'16 & \ding{55} & \ding{55} & \ding{51} &  85.39\\
TF \cite{tf} & CVPR'17 & \ding{55} & \ding{55} & \ding{51} &  80.69\\
TCN \cite{tcn} & CVPRW'17 & \ding{55} & \ding{55} & \ding{51} & 78.57 \\
ST-GCN \cite{stgcn} & AAAI'18 & \ding{55} & \ding{55} & \ding{51} &  81.30\\
H+O \cite{ho} & CVPR'19 & \ding{51} & \ding{55} & \ding{55} &  82.43\\
TTN \cite{ttn} & CVPR'19 & \ding{55} & \ding{55} & \ding{51} & 83.10 \\
\textcolor{black}{DARTS} \cite{darts} & ICLR'19 & \ding{55} & \ding{55} & \ding{51} & 74.26 \\
\textcolor{black}{FairDARTS} \cite{faird} & ECCV'20 & \ding{55} & \ding{55} & \ding{51} & 76.87 \\
LEML \cite{leml} & ICML'15 & \ding{55} & \ding{55} & \ding{51} & 79.48 \\
SPDML\cite{pml}  & TPAMI'18 & \ding{55} & \ding{55} & \ding{51} &  76.52\\
MRMML \cite{mrmml} & TBD'22 & \ding{55} & \ding{55} & \ding{51} &  83.33\\
GEMKML \cite{gemkml} & TMM'21 & \ding{55} & \ding{55} & \ding{51} &  81.75\\
SymNet \cite{symnet} & TNNLS'22 & \ding{55} & \ding{55} & \ding{51} & {82.96}\\
SPDNet \cite{spdnet} & AAAI'17 & \ding{55} & \ding{55} & \ding{51} & 86.26\\
SPDNetBN \cite{spdnetbn} & NeurIPS'19 & \ding{55} & \ding{55} & \ding{51} & {86.83}\\ 
\hline
{DreamNet-27} &  & \ding{55} & \ding{55} & \ding{51} &{87.78}\\
{DreamNet-47} &  & \ding{55} & \ding{55} & \ding{51} & \textbf{88.64}\\
{DreamNet-92} &  & \ding{55} & \ding{55} & \ding{51} & {88.12}\\
\hline
\end{tabular}}
\end{table}

\subsection{Comparison with State-of-the-art Methods}
For a fair comparison, based on the publicy available source codes, we follow the original recommendations to tune the parameters of each comparative method, and report their best results on all three datasets. 
\textcolor{black}{For DARTS and FairDARTS, we follow the practice of \cite{spdnas} to run their official implementation with default settings by treating the logarithm maps of SPD matrices as the Euclidean data.}
For DeepO2P, its classification accuracy on the AFEW dataset is provided by \cite{spdnet}. 
As LEML and GEMKML are very time-consuming in stepping through all sample pairs, we do not run them on the largest-scale UAV-Human dataset used in this article. \textcolor{black}{Since ManifoldNet requires SPD data points with multiple channels, it is inapplicable to the FPHA and UAV-Human skeleton datasets.}
From Table \ref{tab-au}, it is evident that our 27-layer DreamNet respectively achieves the accuracy of 36.59\% and 44.88\% on the AFEW and UAV-Human datasets, outperforming all the involved competitors. Besides, with the increasing network depth (increasing the number $\rm{E}$ of the cascaded RAEs), the accuracy of the 47/92-layer DreamNets is monotonically improving. 
Here, we also select some popular action recognition models for better comparison on the FPHA dataset. Table \ref{tab-fpha} shows that our 27/47/92-layer DreamNets are the best performers for the hand action recognition task. These observations confirm that our deep embedding learning mechanism developed on the SPD manifolds yield useful geometric features for better visual scene parsing.

\section{Conclusion}
In this paper, we proposed an effective methodology for increasing the depth of SPD neural networks without destroying the geometric information conveyed by image data.
This is achieved by proposing a novel cascading network architecture with multiple Riemannian autoencoder learning stages appended to the backbone SPD network to enrich the deep layers of structured representations. 
Thanks to the insertion of innovative residual-like blocks via shortcut connections, a better incremental learning of residual structural details can be facilitated.
The experimental results suggest that our network is an effective solution against the geometric information degradation problem, with 
favourable performance compared to the state-of-the-art methods. \textcolor{black}{
For future work, we plan to develop an adaptive criterion that would enable an automatic assessment of the relative significance of the generated feature maps. This would facilitate the use of a neural architecture search (NAS) technique to adapt the proposed network to different computer vision tasks}.

\ifCLASSOPTIONcaptionsoff
  \newpage
\fi



%

\bibliographystyle{IEEEtran}
\bibliography{egbib}

%




\end{document}